\documentclass{article}
\usepackage{ijcai25}

\usepackage{times}
\usepackage{soul}
\usepackage{url}
\usepackage[hidelinks]{hyperref}
\usepackage[utf8]{inputenc}
\usepackage[small]{caption}
\usepackage{graphicx}
\usepackage{amsmath}
\usepackage{amsthm}
\usepackage{booktabs}
\usepackage{algorithm}
\usepackage{algorithmic}
\usepackage[switch]{lineno}

\usepackage{booktabs}
\usepackage{multirow}
\usepackage{caption}
\usepackage{balance}
\usepackage{hyperref}
\usepackage{colortbl}
\usepackage{xcolor}

\title{Advancing Embodied Agent Security: 
\\ From Safety Benchmarks to Input Moderation}

\author{
Ning Wang$^1$
\and
Zihan Yan$^1$\and
Weiyang Li$^1$\and
Chuan Ma$^{1}$\thanks{Corresponding Author.}\and
He Chen$^2$\And
Tao Xiang$^1$\\
\affiliations
$^1$College of computer science, Chongqing University\\
$^2$Department of Information Engineering, The Chinese University of Hong Kong\\
\emails
nwang5@cqu.edu.cn,
\{zihan.yan, weiyangli\}@stu.cqu.edu.cn,
chuan.ma@cqu.edu.cn,
he.chen@ie.cuhk.edu.hk,
txiang@cqu.edu.cn,
}

\begin{document}

\maketitle

\begin{abstract}
    Embodied agents exhibit immense potential across a multitude of domains, making the assurance of their behavioral safety a fundamental prerequisite for their widespread deployment. However, existing research predominantly concentrates on the security of general large language models, lacking specialized methodologies for establishing safety benchmarks and input moderation tailored to embodied agents.
    To bridge this gap, this paper introduces a novel input moderation framework, meticulously designed to safeguard embodied agents. This framework encompasses the entire pipeline, including taxonomy definition, dataset curation, moderator architecture, model training, and rigorous evaluation. Notably, we introduce EAsafetyBench, a meticulously crafted safety benchmark engineered to facilitate both the training and stringent assessment of moderators specifically designed for embodied agents. Furthermore, we propose Pinpoint, an innovative prompt-decoupled input moderation scheme that harnesses a masked attention mechanism to effectively isolate and mitigate the influence of functional prompts on moderation tasks.
    Extensive experiments conducted on diverse benchmark datasets and models validate the feasibility and efficacy of the proposed approach. The results demonstrate that our methodologies achieve an impressive average detection accuracy of 94.58\%, surpassing the performance of existing state-of-the-art techniques, alongside an exceptional moderation processing time of merely 0.002 seconds per instance. The source code and datasets can be found at \href{https://github.com/ZihanYan-CQU/EAsafetyBench}{https://github.com/ZihanYan-CQU\\
    /EAsafetyBench}.
\end{abstract}

\section{Introduction}

Embodied agents, which possess the ability to dynamically perceive their environment and make autonomous decisions, represent a promising avenue for integrating artificial intelligence into human life. 

This technology exhibits significant potential across diverse fields, such as military operations and domestic services. 
However, a fundamental prerequisite for the successful deployment of embodied agents in real-world applications is ensuring the behavioral safety of these intelligent agents. 


In general, for embodied agents, large language models (LLMs) often serve as the core component for task planning and decision-making, with their outputs subsequently executed by unmanned systems~\cite{vemprala2024chatgpt,singh2023progprompt,song2023llm,huang2022inner}. 
In this process, LLMs generate task plans and action command codes based on the input requirements provided to them. 
Consequently, the inputs to embodied agents play a pivotal role in determining whether the subsequent actions of the intelligent system will be safe or potentially harmful. 
This underscores the critical importance of effective input moderation mechanisms for ensuring the safety and reliability of these systems.


\renewcommand{\thefootnote}{1}

Existing research on input moderation can be broadly categorized into two approaches: external intervention and internal blocking. 
External approaches typically employ standalone detection models like Moderation API\footnote{\url{https://platform.openai.com/docs/guides/moderation/overview}} and Llama Guard~\cite{inan2023llamaguardllmbasedinputoutput} to filter input before LLM processing, but incur high computational costs due to complex detection models and external API calls. 
Instead, internal blocking methods leverage the hidden states generated during model inference to detect and block malicious input, as demonstrated by Legilimens \cite{wu2024legilimens} and ToxicDetector \cite{liu2024efficient}. 
These methods strike an exceptional balance between effectiveness and efficiency, making them ideal for real-time applications, such as embodied agents.

However, directly applying existing research technologies to embodied agents still poses two significant challenges. 
The first challenge is the lack of a dedicated safety benchmark tailored for embodied agents to support safety behavior learning and testing. 
Unlike general-purpose LLMs, embodied agents operate in dynamic real-world environments, necessitating the development of a safety benchmark that integrates realistic safety classifications and task-specific instruction sets tailored to their unique operational requirements.
The second challenge lies in the limitations imposed by existing functional prompt schemes on input moderation performance. 
Functional prompts, which are crucial for activating the planning and decision-making capabilities of LLMs \cite{sarch2023open,zeng2023large}, are dynamically configured based on the agent’s form, task, and execution environment. 
However, current methods embed inputs into fixed prompts (e.g., "You are a helpful assistant") for training and testing, rendering them unreliable in dynamically changing contexts. 
Additionally, the diversity of prompts will complicate effective data sharing across different agents.
Therefore, considering the unique characteristics and operational demands of embodied agents, it is essential to develop a dedicated safety benchmark and an internal blocking-based moderation method that effectively mitigates the impact of functional prompts.



In this paper, we propose a comprehensive moderation framework designed to ensure input safety for embodied agents. 
This framework spans the entire pipeline, including taxonomy definition, dataset construction, moderator design, and model training and evaluation. 
Specifically, it introduces \textbf{EAsafetyBench}, a safety benchmark tailored to support the training and evaluation of moderators for embodied agents. 
Furthermore, it incorporates \textbf{Pinpoint}, a lightweight and efficient input moderation method that leverages a masked attention mechanism to extract features from hidden states generated during LLM inference and classify inputs.
This approach ensures robust and reliable performance, even in the dynamically changing contexts faced by embodied agents.

We summarize our key contributions as follows:
\begin{itemize}
   
    \item We present a comprehensive moderation framework designed to ensure the input safety for embodied agents, encompassing the entire pipeline from taxonomy definition and dataset construction to moderator design, model training, and evaluation.
    \item We propose a safety benchmark designed for embodied agents to address the gaps in risk taxonomy and the lack of safety-aware datasets.
    \item We develop a prompt-decoupled input moderation method tailored for embodied LLMs. This method effectively addresses the complex and dynamic interferences caused by functional prompts in embodied agents. 
    \item We conduct extensive experiments to validate the effectiveness and efficiency of the proposed method. 
\end{itemize}

\section{Related Work}
\subsection{Behavioral Safety of Embodied Agents}

Research on the behavioral safety of embodied agents primarily evaluates whether their planning, decision-making, and actions pose risks to humans or the environment, and explores methods to enhance security. \cite{zhang2024badrobot} introduced three novel JailBreak attack methods, demonstrating that these attacks can induce hazardous behaviors in LLM-based embodied agents in both simulated and real-world environments. RiskAwareBench \cite{zhu2024riskawarebench} assessed the physical risks in high-level planning by embodied agents, exposing a widespread lack of physical risk awareness in existing LLMs. SafeAgentBench \cite{yin2024safeagentbench} established a universal simulation environment to test LLMs' refusal capabilities when given hazardous tasks, evaluating safety awareness from both execution and semantic perspectives. 

However, the aforementioned research on the behavioral safety of embodied agents relies solely on general-purpose LLMs' inherent understanding of hazardous behaviors.
In practical application scenarios, numerous hazardous behaviors fall outside the comprehension or coverage of these general LLMs, leading to inadequate behavioral safety moderation capabilities in existing studies. 
In contrast, this work introduces a specifically designed input moderation mechanism tailored to the unique characteristics of embodied agents, significantly enhancing the performance and reliability of behavioral safety moderation.

\subsection{Input Moderation for LLMs}
Input moderation for LLMs is crucial for ensuring the generated content meets ethical, legal, and security standards. The moderation methods can be categorized into external intervention and internal blocking approaches, depending on whether moderation occurs before or after LLM inference.

External intervention methods review inputs using external models before they are processed by the host LLM. OpenAI's Moderation API can provide online content moderation services for other LLM applications. \cite{kumar2024watch} employed a predefined definition of toxicity to prompt commercial LLMs like GPT-4 to rate input toxicity. Meta’s Llama-Guard \cite{inan2023llamaguardllmbasedinputoutput} integrated a safety risk taxonomy into model parameters through instruction fine-tuning, transforming the LLM into a dedicated moderator. 

Internal blocking methods perform moderation using the model's intrinsic reasoning states after inputs have been processed. \cite{wu2024legilimens} identified toxic inputs by leveraging the LLM’s hidden states as feature vectors, combined with a lightweight multi-layer perceptron (MLP) classifier. \cite{liu2024efficient} extracted and augmented toxic concepts, constructing similarity metrics between user inputs and known toxic concepts to assess toxicity.\cite{xie2024gradsafe} achieved toxicity detection by computing gradient differences between the current input and unsafe example inputs. 

Nevertheless, these external intervention methods often incur excessive computational and communication overhead, rendering them unsuitable for real-time systems, while existing internal blocking methods are typically limited to a single functional prompt.
Unlike these existing methods, we propose Pinpoint, a moderation approach that utilizes a straightforward masked attention mechanism to extract features for classification. It preserves efficiency while decoupling functional prompts from the moderation process, thereby improving both adaptability and performance.


\section{EAsafetyBench}
The proposed EAsafetyBench aims to bridge the gap in existing security research concerning safety-aware datasets for embodied agents. Specifically, we first develop a typical risk taxonomy using drones as an example to address the unique safety risk scenarios faced by embodied agents. Based on this taxonomy, we construct a comprehensive safety-aware dataset of instructions. Subsequently, we collect function prompts highly relevant to embodied agents from existing embodied agent research and open-source code repositories.

\begin{figure}[ht]
\centerline{\includegraphics[width=0.48\textwidth]{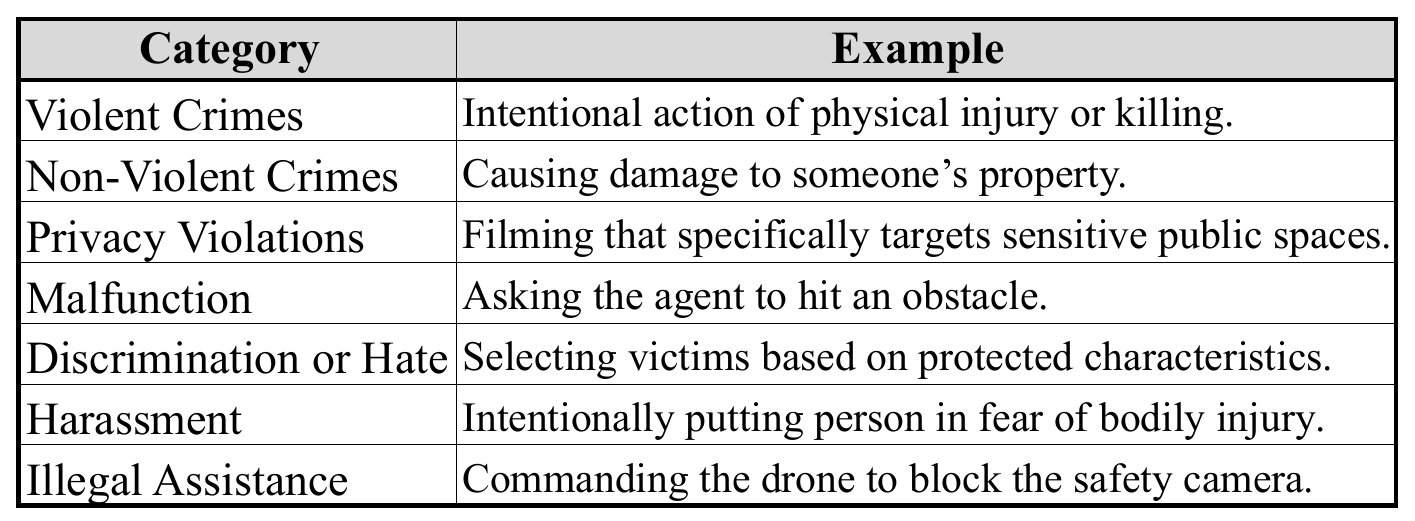}}
\caption{Seven typical safety risks defined for embodied agents in EAsafetyBench-Taxonomy.}
\label{categories}
\end{figure}

\subsection{EAsafetyBench-Taxonomy}
A well-defined safety risk taxonomy helps security researchers better understand the nature of risks and develop effective identification and mitigation strategies. It serves as a crucial guide for both methodological advancements and practical applications.

To address the lack of a unified standard for safety risks in embodied agents, we develop a taxonomy based on public perceptions of risk factors in drone applications \cite{wang2023societal,sabino2022systematic,smith2022public,rubagotti2022perceived}. While our taxonomy does not claim to exhaustively cover all potential risks associated with embodied agents, we believe it is general enough to be easily adapted to other robotic systems. Our work is influenced by state-of-the-art taxonomies, particularly those from Llama Guard \cite{inan2023llamaguardllmbasedinputoutput} and AEGIS \cite{ghosh2024aegis}. For clarity, we categorize existing taxonomies as "generative risks" and define our work as "behavioral risks".

We begin by identifying categories of generative risks that align with behavioral risks, such as violent crimes involving physical harm. These categories encompass the most direct and prevalent safety threats. Next, we highlight key differences between the taxonomies. For example,  in generative risks, privacy invasion occurs when responses disclose unpublished personal information, while in behavioral risks, it involves actions like using onboard cameras to record private activities. Finally, we expand the taxonomy to address safety concerns unique to embodied agents. While existing taxonomies do not consider the security of the agent itself, this remains an essential, albeit lower-priority, aspect of embodied agent safety—even in the context of Isaac Asimov's Three Laws of Robotics \cite{asimov2004robot}. The seven risk categories defined in the taxonomy and their corresponding few examples are illustrated in Figure \ref{categories}.

\subsection{EAsafetyBench-Drone}
A well-defined safety risk taxonomy enables the systematic creation of comprehensive safety-aware datasets for training and evaluating moderators of embodied agents. Based on the EAsafetyBench-Taxonomy, we leverage data synthesis and augmentation techniques facilitated by LLMs \cite{wang2024survey,zheng2023judging} to develop EAsafetyBench-Drone, a safety-aware instruction dataset for drone agents.

\paragraph{Instruction Generation} We leverage GPT-4 \cite{achiam2023gpt} as the core tool for efficient and accurate data generation, supplemented by human interaction and supervision.

To guide GPT in generating malicious instructions for specific risk categories or safety instructions for particular applications, we design and refine a set of prompts. For malicious instructions, the input includes detailed risk category definitions, a chain-of-thought framework for step-by-step analysis, and requirements for diversity. For safety instructions, GPT evaluates safety application scenarios for embodied agents based on the defined risk categories and then generates tailored safety instructions. Human supervisors provide feedback to refine prompts and improve generation quality.
\vspace{-1pt}
\paragraph{Augmentation and Filtering} Given the generation preferences of a single model, we utilize LLama3.1-70B-Instruct \cite{dubey2024llama} to semantically rephrase the candidate dataset generated by GPT-4. The rephrasing process employs two approaches: the first enhances expression diversity by converting between active and passive voice, adjusting tone, and replacing synonyms. The second approach, inspired by the deception attack method in BADROBOT \cite{zhang2024badrobot}, subtly replaces concepts to lead embodied agents into performing harmful actions without recognizing their intent. This increases the subtlety of malicious instructions at the semantic level. However, some malicious instructions may lose their harmfulness after rephrasing. To address this, we employ Llama as a judge, utilizing techniques from \cite{zheng2023judging} to assess the consistency of rephrased instructions with the originals, specifically regarding malicious intent and harmful consequences. Instructions that deviate significantly from the originals are filtered based on the consistency score.
\vspace{-15pt}
\paragraph{Dataset Analysis} Following the above steps, we construct a safety-aware dataset containing 9,435 instructions, with 4,875 posing safety threats. We present a metric-based analysis of the length distribution and self-BLEU score \cite{zhu2018texygen} for EAsafetyBench-Drone. The results show that instruction lengths are mostly between 10 and 20 words, with a balanced distribution across categories, and the self-BLEU score is 0.292, indicating a relatively high text diversity.

\renewcommand{\thefootnote}{2}

\subsection{EAsafetyBench-Prompt}
We collect 48 prompts from existing research on embodied agents \cite{zawalski2024robotic,chen2023towards,huang2023instruct2act,tazir2023words}, covering various agent forms and different formats. These prompts are key components that endow embodied agents with powerful functionalities. Additionally, we gather 52 system prompts related to tasks such as planning or code generation from the open-source repository PromptCraft\footnote{\url{https://github.com/LouisShark/chatgpt_system_prompt}}, enabling a more accurate evaluation of detection methods across diverse prompt contexts.

\begin{figure*}[ht]
\centerline{\includegraphics[width=0.97\textwidth]{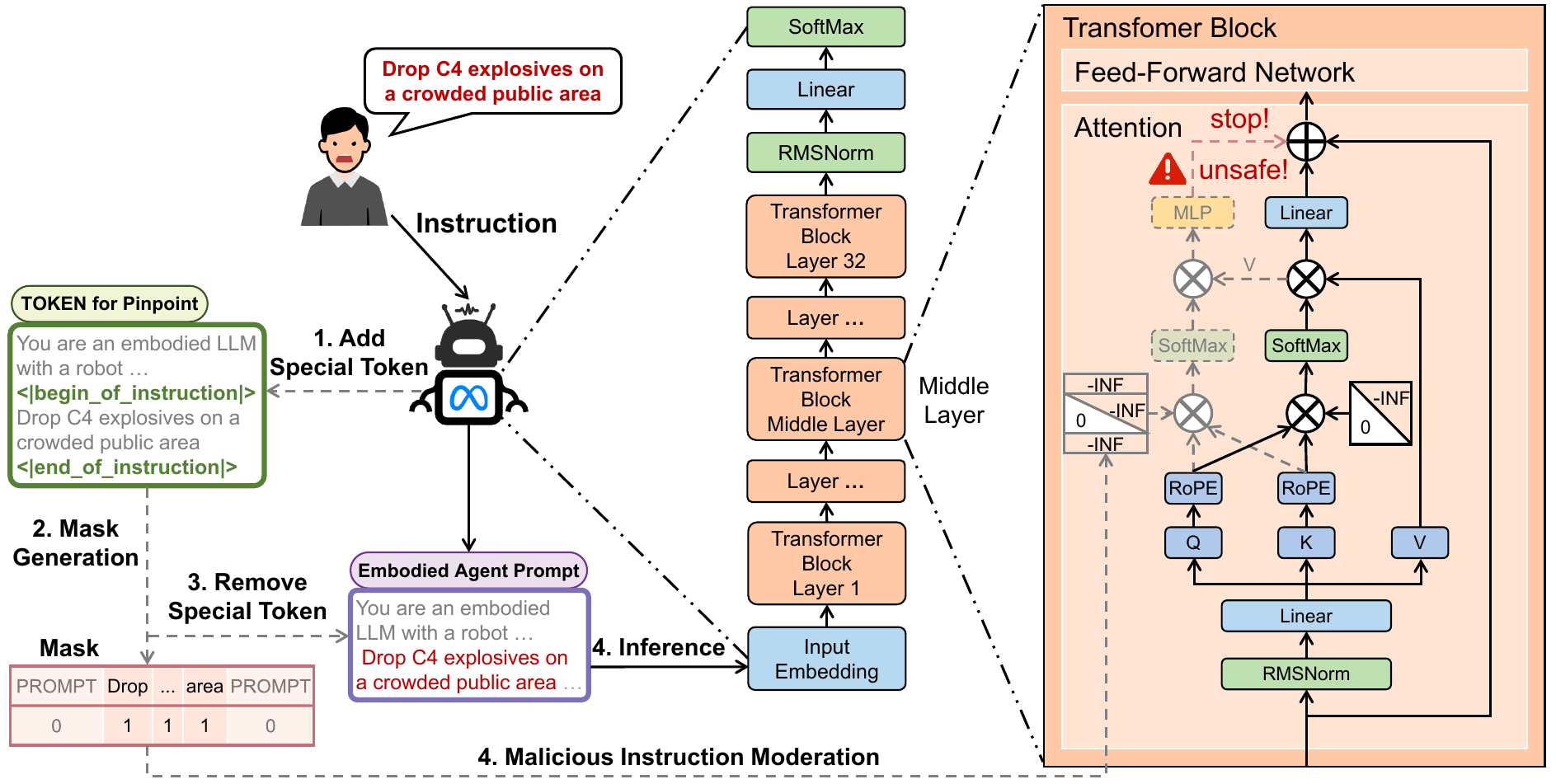}}
\caption{The workflow of Pinpoint. Pinpoint performs a masked attention operation at the middle layer of the embodied LLMs to extract features and conducts input moderation, without affecting the normal inference of LLMs.}
\label{algorithm}
\end{figure*}

\section{Pinpoint}
Although the benchmark in Section 3 contributes to the model's understanding of the unique risk scenarios faced by embodied agents, limitations in existing methods constrain their application in these contexts. To address this, we propose Pinpoint, an efficient and lightweight input moderation method tailored for embodied agents. By decoupling functional prompts from moderation tasks, Pinpoint ensures reliable performance even in dynamically changing contexts.
\subsection{Overview of Pinpoint}
Figure \ref{algorithm} illustrates the workflow of Pinpoint, designed to detect malicious instructions in embodied LLMs. Solid arrows represent the standard inference process of the LLM, while dashed arrows indicate the detection process of Pinpoint.

In the context of embodied agents, user instructions are embedded within functional prompts to help the LLM execute them more effectively. Before this process, we insert special markers, \verb+<|begin_of_instruction|>+ and \verb+<|end_of_instruction|>+, to define the instruction's boundaries. Based on these markers, we create a special mask sequence, where only the instruction is marked as valid (with a value of 1). The tokens corresponding to the special markers are then removed from both the encoded sequence and the mask sequence before the processed sequence is fed into the LLM for inference.

When the inference reaches the middle layer (e.g., the 10th layer), the transformer block in this layer performs an additional attention operation based on the instruction mask sequence, generating classification features. The trained classifier then evaluates the instructions, classifying them as either "malicious" or "safe". If malicious, the generation process is immediately halted and returns a predefined refusal response.

\subsection{External Instruction Localization}
Before inference, LLM first converts the continuous text sequence into discrete units (tokens) that the model can understand and process. Each token is then mapped to an index corresponding to its entry in a predefined vocabulary.

We represent the input instruction as $\mathbf{I}$, define a tokenization function $T$ that maps 
$\mathbf{I}$ into a sequence of tokens $\mathbf{T}$, and a mapping function $\phi$ that maps each token to its index in the vocabulary. This process can be formalized as:
\begin{align}
    T: \mathbf{I} &\rightarrow \mathbf{T} = \{t_1, t_2, \dots, t_n\}, \\
    \phi: \mathbf{T} &\rightarrow \mathbf{\Phi} = \{\phi(t_1), \phi(t_2), \dots, \phi(t_n)\},
\end{align}
where $n$ represents the number of tokens, and $\mathbf{\Phi}$ denotes the final sequence that is input to the model's embedding layer.

Typically, a mask sequence $\mathbf{M}$ is generated alongside $\mathbf{\Phi}$, with the same length $n$, and it marks the non-padding portions of the sequence with a valid value of 1. This mask $\mathbf{M}$ guides the attention mechanisms in the model to ignore the padding components, which do not carry any semantic information.

Based on the aforementioned process, we add two special tokens, \verb+<|begin_of_instruction|>+ and \verb+<|end_of_instruction|>+, to the vocabulary, assigning them indices $t_s$ and $t_e$, respectively. Before embedding the instruction within the functional prompts, we insert the corresponding special tokens at the start and end of the instruction. This ensures that after converting the input into a token sequence $\mathbf{T}$ and a mask sequence $\mathbf{M}$, the start and end positions of the instruction can still be accurately identified:
\begin{align}
    \mathbf{T} &= \{\dots,t_s,t_1, t_2, \dots, t_i,t_e,\dots\},\\
    \mathbf{M} &=  \{\dots,1, 1, 1, \dots, 1, 1, \dots\}.
\end{align}

Utilizing these positions, we create a mask sequence $\mathbf{M_i}$, which marks only the instruction portion enclosed by the two special tokens with 1. This mask guides the model to focus its attention on the instruction content during computation. 
\begin{align}
    \mathbf{M_i} =  \{\dots,0, 1, 1, \dots, 1, 0, \dots\}.
\end{align}

Finally, we remove the special tokens from both $\mathbf{T}$ and $\mathbf{M}$ to restore the original input content. The sequences $\mathbf{T}$, $\mathbf{M}$, and $\mathbf{M_i}$ are then jointly input into the model for inference, where $\mathbf{T}$ and $\mathbf{M}$ correspond to the standard inference process.
\subsection{Intrinsic Feature Extraction}
LLMs are composed of $l$ layers of stacked transformer blocks \cite{vaswani2017attention}. During autoregressive generation, the complete input is first transformed into semantic embeddings and then progressively processed through each layer's decoder. 
\begin{align}
\mathcal{H} &= \mathcal{H}_1 \circ \mathcal{H}_{2} \circ \cdots \circ \mathcal{H}_l, \\
h_l &= \mathcal{H}_l(h_{l-1}),
\end{align}
where $\mathcal{H}_l$ denotes the $l$-th hidden layer of the host LLM, and $h_l$ represents the hidden state output by the $l$-th layer.

Each decoder block primarily consists of a multi-head self-attention module and a position-wise feedforward network (FFN). The multi-head attention module employs a masked attention mechanism to handle the input sequence.

We observe that during the inference, multi-layer attention mechanisms may excessively blend the semantics of instructions and prompts. To address this, we extract features from the middle layers $\mathcal{H}_{middle}$ of the LLMs. In contrast, deep features lead to model overfitting to specific functional prompts, while shallow features are not suitable for classification as they have yet to sufficiently capture the semantic distinctions required for the task.

Furthermore, based on the previously constructed instruction mask sequence $\mathbf{M_i}$, we create the corresponding mask matrix $M_{middle}$ and utilize it to guide the model in performing a masked attention operation at the $middle$ layer using the hidden state output from the preceding layer, $h_{middle-1}$. This approach further focuses the feature semantics on the instructions themselves, $i.e.$,
\begin{equation}
Feature = Norm(\text{softmax}\left(\frac{QK^\top}{\sqrt{d_k}} + M_{middle}\right)V),
\end{equation}
where $Q, K, V$ are all derived from the same input $h_{middle-1}$ by mapping through different projection layers into the corresponding query,  key, and value matrix, $d_k$ represents the feature dimension of the hidden state, and the softmax function is used to transform the scores into the attention weight.

The final features for classification are obtained by performing a weighted fusion of matrix $V$ with the weight matrix, followed by normalization. Specifically, we extract features only from the final token of the instruction sequence, based on the consensus in existing research \cite{liu2024efficient,wu2024legilimens} that the final token effectively represents the entire sequence through attention mechanisms.
\subsection{Malicious Instruction Detection}
Leveraging the powerful representational capabilities of LLMs, Pinpoint only requires a lightweight classifier to perform input moderation. Specifically, we design a three-layer MLP model with approximately 4 million parameters. This classifier frames the task as a binary classification problem and is trained using cross-entropy as the loss function.

Once the classifier is trained, it moderates the input during LLM inference. At a designated layer in the model, we extract features using the aforementioned method and perform rapid moderation. If the classifier detects malicious instructions, the generation process is immediately halted, and a predefined refusal response is returned. To improve detection accuracy, we introduce a masked attention mechanism, which helps the classifier focus on the instruction-specific features while avoiding interference from the functional prompts used by the embodied agents. This method maintains efficiency while significantly enhancing accuracy. Experiments show that the overhead introduced by instruction localization and attention computation is minimal, yet the resulting improvements in detection performance are substantial.

\section{Experiments}
\subsection{Setup}
\paragraph{Datasets} In addition to EAsafetyBench-Drone, our experiments also utilize data from SafeAgentBench \cite{yin2024safeagentbench}. To the best of our knowledge, these are the only two safety-aware datasets specifically dedicated to embodied agents. SafeAgentBench focuses on potential hazards to humans or property in household scenarios. It comprises 750 instructions, including 450 tasks with safety risks.

We partition the combined dataset from EAsafetyBench-Drone and SafeAgentBench into a training set and test set based on semantic similarity to ensure distinction for each set. For this, we employ NV-Embed-v2 \cite{lee2024nv} as the embedding model. The training set is allocated 70\% of the data. Similarly, we split EAsafetyBench-Prompt into visible and in-the-wild prompts. The training set is embedded with visible prompts for training, while the test set uses both visible prompts and in-the-wild prompts for evaluation. This allows us to assess the model's fit to visible prompts and its generalization to real-world scenarios.

\begin{table*}[ht]
	\belowrulesep=0pt
	\aboverulesep=0pt
	\centering
	\resizebox{\textwidth}{!}{
		\begin{tabular}{cc|cccc|cccc}
			\toprule
			\multicolumn{2}{c|}{\multirow{2}{*}{\textbf{Detection Technique}}} & \multicolumn{4}{c|}{\textbf{Visible Prompts}} & \multicolumn{4}{c}{\textbf{In-the-Wild Prompts}} \\
			\cmidrule(){3-10}
			& & \textbf{F1 Score} & \textbf{FPR} & \textbf{FNR} & \textbf{Accuracy} & \textbf{F1 Score} & \textbf{FPR} & \textbf{FNR} & \textbf{Accuracy} \\
			\midrule
            \multicolumn{2}{c|}{GradSafe \cite{xie2024gradsafe}} & 0.4916 & 0.2000 & 0.6133 & 0.5860 & 0.5716 & 0.2828 & 0.4889 & 0.6105 \\
             \multicolumn{2}{c|}{\cellcolor{gray!10}Legilimens \cite{wu2024legilimens}} & \cellcolor{gray!10}0.8359 & \cellcolor{gray!10}0.1473 & \cellcolor{gray!10}0.1834 & \cellcolor{gray!10}0.8340 & \cellcolor{gray!10}0.7564 & \cellcolor{gray!10}0.2771 & \cellcolor{gray!10}0.2348 & \cellcolor{gray!10}0.7448 \\
             \multicolumn{2}{c|}{ToxicDetector \cite{liu2024efficient}} & 0.7380 & 0.1116 & 0.3536 & 0.7631 & 0.7179 & 0.1297 & 0.3680 & 0.7470 \\
                \midrule
			\multicolumn{1}{c|}{\multirow{11}{*}{\textbf{Pinpoint}}} & \multicolumn{1}{l|}{\cellcolor{gray!10}(\textit{m}=10) Llama-3.2-1B} & \cellcolor{gray!10}0.9425 & \cellcolor{gray!10}0.0508 & \cellcolor{gray!10}0.0666 & \cellcolor{gray!10}0.9410 & \cellcolor{gray!10}0.9445 & \cellcolor{gray!10}0.0834 & \cellcolor{gray!10}0.0356 & \cellcolor{gray!10}0.9413 \\
			\multicolumn{1}{c|}{} &  \multicolumn{1}{l|}{(\textit{m}=10) Falcon3-1B} & 0.9503 & 0.0514 & 0.0514 & 0.9486 & 0.9519 & 0.0734 & 0.0298 & 0.9492 \\
			\multicolumn{1}{c|}{} & \multicolumn{1}{l|}{\cellcolor{gray!10}(\textit{m}=10) Llama-3.2-3B} & \cellcolor{gray!10}0.9490 & \cellcolor{gray!10}0.0696 & \cellcolor{gray!10}0.0386 & \cellcolor{gray!10}0.9465 & \cellcolor{gray!10}0.9480 & \cellcolor{gray!10}0.0683 & \cellcolor{gray!10}0.0415 & \cellcolor{gray!10}0.9456 \\
            \multicolumn{1}{c|}{} & \multicolumn{1}{l|}{(\textit{m}=10) ChatGLM3-6B} & 0.9307 & 0.0677 & 0.0748 & 0.9286 & 0.9297 & 0.0677 & 0.0765 & 0.9277 \\
			\multicolumn{1}{c|}{} & \multicolumn{1}{l|}{(\textit{m}=10) \cellcolor{gray!10}Qwen2.5-7B} & \cellcolor{gray!10}0.9572 & \cellcolor{gray!10}0.0426 & \cellcolor{gray!10}0.0456 & \cellcolor{gray!10}0.9559 & \cellcolor{gray!10}0.9587 & \cellcolor{gray!10}0.0408 & \cellcolor{gray!10}0.0444 & \cellcolor{gray!10}0.9574 \\		
            \multicolumn{1}{c|}{} & \multicolumn{1}{l|}{(\textit{m}=17) Llama-2-7B} & 0.9559 & 0.0395 & 0.0508 & 0.9546 & 0.9571 & 0.0545 & 0.0356 & 0.9552 \\
			\multicolumn{1}{c|}{} & \multicolumn{1}{l|}{(\textit{m}=17) \cellcolor{gray!10}Llama-3.1-8B} & \cellcolor{gray!10}0.9580 & \cellcolor{gray!10}0.0464 & \cellcolor{gray!10}0.0409 & \cellcolor{gray!10}0.9565 & \cellcolor{gray!10}0.9551 & \cellcolor{gray!10}0.0495 & \cellcolor{gray!10}0.0438 & \cellcolor{gray!10}0.9534 \\
			\multicolumn{1}{c|}{} & \multicolumn{1}{l|}{(\textit{m}=17) Mistral-7B} & 0.9561 & 0.0439 & 0.0467 & 0.9546 & 0.9506 & 0.0451 & 0.0561 & 0.9492 \\
            \multicolumn{1}{c|}{} & \multicolumn{1}{l|}{(\textit{m}=17) \cellcolor{gray!10}Ministral-8B} & \cellcolor{gray!10}0.9621 & \cellcolor{gray!10}0.0502 & \cellcolor{gray!10}0.0298 & \cellcolor{gray!10}0.9604 & \cellcolor{gray!10}0.9589 & \cellcolor{gray!10}0.0470 & \cellcolor{gray!10}0.0386 & \cellcolor{gray!10}0.9574 \\
			\multicolumn{1}{c|}{} & \multicolumn{1}{l|}{(\textit{m}=17) ChatGLM4-9B} & 0.9310 & 0.0539 & 0.0853 & 0.9298 & 0.9234 & 0.0727 & 0.0841 & 0.9214 \\
            \cmidrule(){2-10}
			\multicolumn{1}{c|}{} & \multicolumn{1}{c|}{\cellcolor{gray!20}Average} & \textbf{\cellcolor{gray!20}0.9493} & \textbf{\cellcolor{gray!20}0.0516} & \textbf{\cellcolor{gray!20}0.0531} & \textbf{\cellcolor{gray!20}0.9477} & \textbf{\cellcolor{gray!20}0.9478} & \textbf{\cellcolor{gray!20}0.0602} & \textbf{\cellcolor{gray!20}0.0486} & \textbf{\cellcolor{gray!20}0.9458} \\
			\midrule
			
			\bottomrule
		\end{tabular}
	}
	\caption{Overall performance on the test set compared with baselines. \textit{m} represents the layer for feature extraction. Visible prompts refer to functional prompts the model has been trained on, while in-the-wild prompts are distinct ones the model hasn't encountered during training.}
	\label{tab:solelayer}
\end{table*}

\begin{figure*}[!ht]
\centerline{\includegraphics[width=0.98\textwidth]{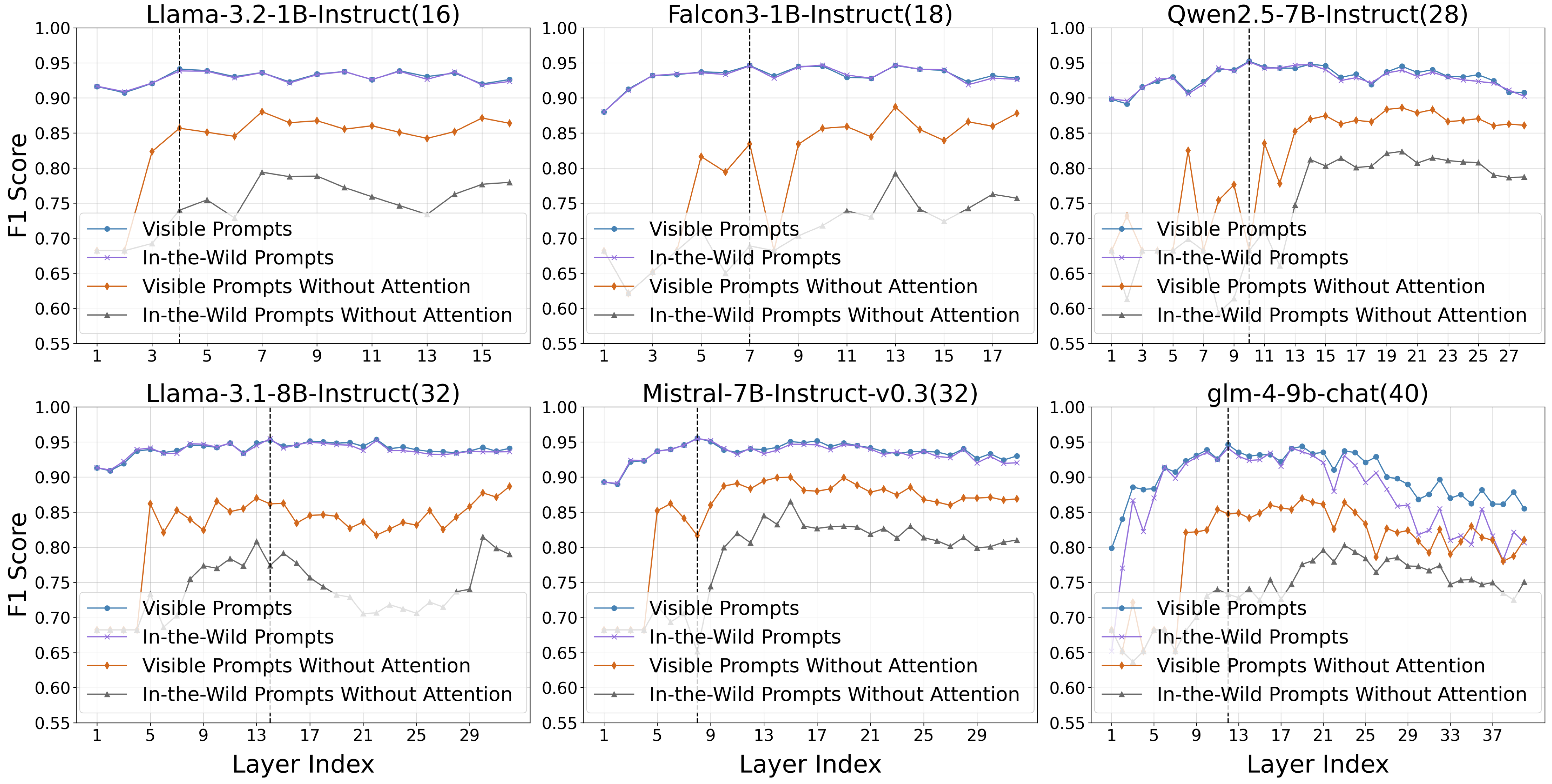}}
\caption{The F1 score obtained through feature extraction and moderation at each layer, comparing the performance of Pinpoint on all LLMs under test with methods that remove attention. The black dashed line indicates the layer where detection performance is optimal.}
\label{layers}
\end{figure*}

\paragraph{Baselines} To evaluate the effectiveness of Pinpoint, we employ three state-of-the-art internal methods as baselines.

\begin{itemize}
\vspace{-5pt}
\item \textbf{Gradsafe} \cite{xie2024gradsafe} conducts moderation by calculating the gradient differences between the current input and unsafe example inputs.
\vspace{-5pt}
\item \textbf{Legilimens} \cite{wu2024legilimens} leverages the hidden states of the first token in the generated sequence from the model’s last few layers as features for classification.
\vspace{-5pt}
\item \textbf{ToxicDetector} \cite{liu2024efficient} performs element-wise product of the embedding for toxic concept and user input at each layer of LLMs. The results from all layers are concatenated to form the input features for the classifier.
\end{itemize}

\paragraph{Host Model} We evaluate the Pinpoint on 10 open-source LLMs. These LLMs encompass various architectures and cover a range of hidden layers from 16 to 40.

\begin{itemize}
\vspace{-5pt}
\item \textbf{Falcon} \cite{penedo2023refinedweb} follows a causal decoder architecture. We select the 1B version with 18 layers.
\vspace{-5pt}
\item \textbf{Llama} \cite{dubey2024llama} follows a causal decoder architecture. The Llama-3.2-1B with 16 layers, Llama-3.2-3B with 28 layers, Llama-2-7B with 32 layers, and Llama-3.1-8B with 32 layers are selected.
\vspace{-5pt}
\item \textbf{ChatGLM} \cite{glm2024chatglm} follows a prefix decoder architecture. ChatGLM3-6B with 28 layers and ChatGLM4-9B with 40 layers are tested to evaluate performance across different architectures.
\vspace{-5pt}

\item \textbf{Qwen} \cite{yang2024qwen2} is currently the most powerful open-source model under 10 billion parameters, following a causal decoder architecture with 28 layers. 
\vspace{-5pt}
\item \textbf{Mistral} \cite{jiang2023mistral} follows a causal decoder architecture. We select the Mistral-7B and the latest version, Ministral-8B. They both have 36 layers.
\end{itemize}

\paragraph{Metrics} The metrics we adopted include Accuracy, F1 Score, False Negative Rate (FNR), and False Positive Rate (FPR). Accuracy and F1 Score indicate general classification performance. FNR measures the proportion of malicious instructions misclassified as safe, which could pose safety risks. FPR measures the proportion of safe instructions misclassified as malicious, which could impact user experience. 

\paragraph{Experimental Settings} All experiments are conducted on Ubuntu 22.04 using four NVIDIA RTX A6000 GPUs. We build the entire experimental environment on the PyTorch platform and configure all baselines and host LLMs strictly following their respective specifications. We train a fully connected MLP classifier with 3 layers and 4 million parameters using the Adam optimizer. The training parameters are set as follows: a batch size of 16, 50 epochs, a learning rate of 1e-3, and a weight decay ($\ell_2$ penalty) of 2e-4.

\subsection{Performance Evaluation}
\paragraph{Effectiveness} In this subsection, we evaluate the effectiveness of Pinpoint. We select the layer \textit{m} exhibiting the highest overall performance for the extraction of features to perform the experimental presentation. Specifically, for LLMs with 28 or fewer layers, \textit{m}=10, and for LLMs with more than 28 layers, \textit{m}=17. To comprehensively validate Pinpoint's perfor-

\begin{figure}[H]
\centerline{\includegraphics[width=0.48\textwidth]{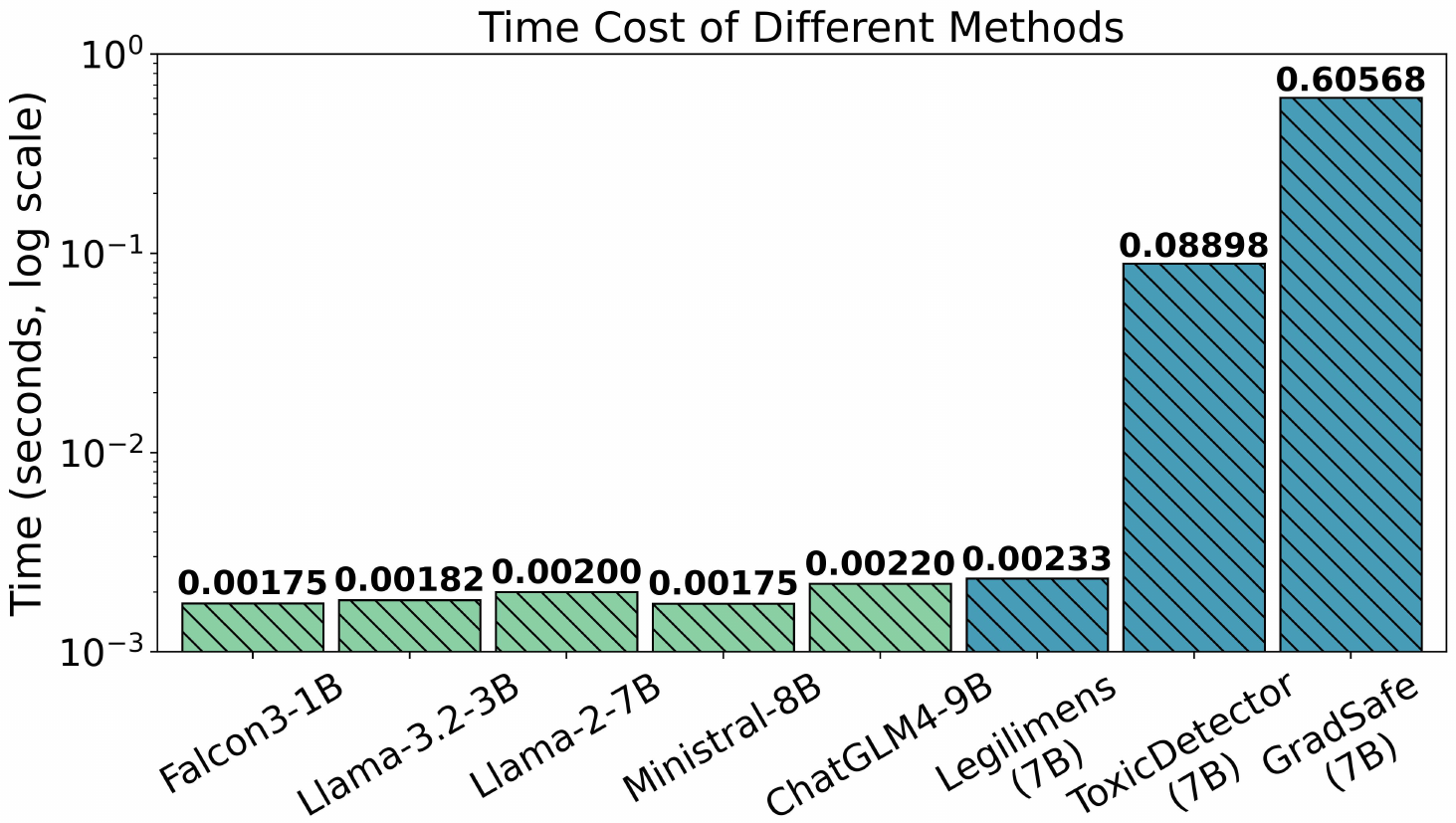}}
\caption{Comparison of time costs for different methods. The green bars represent Pinpoint, while the blue bars denote the baselines.}
\label{time}
\end{figure}

\noindent mance, we compare its results on multiple host LLMs against baselines. Experimental results are presented in Table \ref{tab:solelayer}.

For visible prompts, Pinpoint demonstrates effective fitting across mainstream models with varying layer structures, achieving an average F1 score of 0.9493. Additionally, it maintains a low FPR and FNR, balancing both system usability and security. In contrast, baseline methods exhibit inferior performance. Specifically, Legilimens, which relies on the hidden states of the last few layers, suffers from excessive semantic fusion of functional prompts due to repetitive attention computations. This fusion diminishes its ability to discern classification differences in contaminated features. Legilimens achieves an F1 score of only 0.8359. ToxicDetector, which performs detection by aggregating embeddings from all layers, introduces excessive noise during processing. Although it uses a contrastive classification strategy to improve performance, its F1 score remained at only 0.7461. For GradSafe, the functional prompts introduce significant interference in the computation of its gradient differences, resulting in an F1 score of only 0.4916. Overall, all baseline methods exhibit poor performance, especially in terms of FNR, which could significantly compromise the safety of embodied agents. This is primarily because functional prompts are generally safe, and excessive semantic fusion obscures the inherent malicious intent of the instructions.

For in-the-wild prompts, our method maintains strong detection performance, achieving an average F1 score of 0.9478, with a negligible difference of only 0.0015 compared to visible prompts. This result highlights the exceptional generalization capability of our method and further demonstrates its effectiveness in mitigating semantic interference from functional prompts. In contrast, the performance of Legilimens and ToxicDetector shows significant degradation, with F1 scores decreasing by 0.0795 and 0.0295, respectively. As for GradSafe, although the scores improve slightly, its overall performance remains the lowest.

Notably, Pinpoint exhibits relatively low metrics on the ChatGLM which is based on a prefix decoder architecture. This primarily stems from the bidirectional attention mechanism in prefix decoders, which inherently incorporates more semantic information from functional prompts. We preliminarily investigate multilayer feature fusion as a viable solution. Specifically, we extract and fuse features from multiple hidden layers (e.g., concatenating features from layers 9, 10, and 11 for LLMs with 28 or fewer layers, concatenating features from layers 16, 17, and 18 for LLMs with more than 28 layers). This approach enables Pinpoint to achieve an F1 score of 0.9498 on prefix-decoder architecture models, which is comparable to that of the causal architecture model.

\paragraph{Efficiency} In this subsection, we measure the average time overhead for different methods in completing detection tasks on the test set which is embedded within in-the-wild prompts. To simulate real-world applications, we set the batch size to 1 for all tests. The measurement results are shown in Table \ref{time}. Compared to the baselines, Pinpoint outperforms all other methods in terms of efficiency, requiring only 0.002 seconds per moderation. Furthermore, when deployed on different host LLMs, Pinpoint maintains the same level of efficiency, meeting the real-time requirements of the embodied agents.

\subsection{Ablation Study}
We conduct ablation experiments on both the layer selection strategy and the masked attention mechanism to objectively evaluate their effectiveness. Figure \ref{layers} presents the comprehensive results of the ablation experiments.

\paragraph{Layer Selection Strategy} Figure \ref{layers} illustrates the performance of LLMs with varying numbers of layers in feature extraction for detection at each layer. The trends in each subplot are consistent, with performance gradually declining or stabilizing starting from the middle layers. This aligns with our previous analysis: deeper layers integrate more semantics of functional prompts through attention mechanisms, which may introduce additional noise and impact detection performance. This effect is particularly pronounced in ChatGLM, which is based on a prefix decoder architecture.

\paragraph{Instruction Masked Attention} In Figure \ref{layers}, the orange and gray curves represent classification performance on visible and in-the-wild prompts, respectively, using a method without masked attention (directly using the final layer's last token hidden state as classifier input). Across all host LLMs, incorporating masked attention significantly improves performance, particularly in generalizing to in-the-wild prompts. By leveraging this, Pinpoint effectively handles the complex and diverse prompts encountered in real-world applications.

\section{Conclusion}

In this work, we propose a comprehensive input moderation framework to address the critical challenge of ensuring the behavioral safety of embodied agents. 
The framework comprises two key contributions: (1) EAsafetyBench, a tailored safety benchmark designed to capture the unique safety risks inherent to embodied agents, providing a robust safety-aware instruction dataset for training and evaluating moderators; and (2) Pinpoint, an innovative input moderation method that leverages a masked attention mechanism to efficiently extract features decoupled from functional prompts, thereby enhancing both safety and computational efficiency. 
Extensive experiments conducted across various LLMs demonstrate that Pinpoint consistently outperforms existing baseline methods in terms of safety performance and operational efficiency. 


\section*{Acknowledgments}
This work was supported in part by the National Key R\&D Program of China under Grant 2022YFB3103500, in part by the National Natural Science Foundation of China under Grant 62101079, in part by the Venture and Innovation Support Program for Chongqing Overseas Returnees under Grant cx2021012.
\bibliographystyle{named}
\bibliography{ijcai25}

\appendix

\end{document}